\documentclass{article}

\usepackage{arxiv}

\usepackage[utf8]{inputenc} 
\usepackage[T1]{fontenc}    
\usepackage{hyperref}       
\usepackage{url}            
\usepackage{booktabs}       
\usepackage{amsfonts}       
\usepackage{nicefrac}       
\usepackage{microtype}      
\usepackage{lipsum}
\usepackage{graphicx}
\usepackage{algorithm}
\usepackage{algorithmic}
\usepackage{amsmath}
\usepackage{multirow}

\graphicspath{ {./images/} }

\title{A Novel Algorithm for Personalized Federated Learning: Knowledge Distillation with Weighted Combination Loss}

\author{
Hengrui Hu \\
  Division of Biostatistics\\
  Medical College of Wisconsin\\
  Wisconsin, WI 53226 \\
  \texttt{hhu@mcw.edu} \\
   \And
Anai N. Kothari \\
  Department of Surgery\\
  Medical College of Wisconsin\\
  Wisconsin, WI 53226 \\
  \texttt{akothari@mcw.edu} \\
  \And
Anjishnu Banerjee \\
  Division of Biostatistics\\
  Medical College of Wisconsin\\
  Wisconsin, WI 53226 \\
  \texttt{abanerjee@mcw.edu} \\
}

\begin{document}
\maketitle
\begin{abstract}
Federated learning (FL) offers a privacy-preserving framework for distributed machine learning, enabling collaborative model training across diverse clients without centralizing sensitive data. However, statistical heterogeneity, characterized by non-independent and identically distributed (non-IID) client data, poses significant challenges, leading to model drift and poor generalization. This paper proposes a novel algorithm, pFedKD-WCL (Personalized Federated Knowledge Distillation with Weighted Combination Loss), which integrates knowledge distillation with bi-level optimization to address non-IID challenges. pFedKD-WCL leverages the current global model as a teacher to guide local models, optimizing both global convergence and local personalization efficiently. We evaluate pFedKD-WCL on the MNIST dataset and a synthetic dataset with non-IID partitioning, using multinomial logistic regression and multilayer perceptron models. Experimental results demonstrate that pFedKD-WCL outperforms state-of-the-art algorithms, including FedAvg, FedProx, Per-FedAvg, and pFedMe, in terms of accuracy and convergence speed.
\end{abstract}


\section{Introduction}

Federated learning (FL) has emerged as a transformative approach in distributed machine learning, enabling a diverse set of clients, such as mobile devices, edge nodes, or institutions, to collaboratively train a shared model while preserving data privacy by keeping local datasets on the device \cite{MMRH2017}. This privacy-preserving algorithm framework is particularly valuable in domains such as healthcare, finance, and IoT systems, where sensitive data cannot be centralized \cite{kairouz2019advances, yang2019federated}. However, a significant challenge in FL algorithms is the statistical heterogeneity among clients, often termed non-independent and identically distributed (non-IID) data, where local data distributions differ due to varying user behaviors or contexts \cite{li2019convergence, sattler2019robust}. This diversity causes local models to drift from the global objective, slowing convergence and limiting the global model's ability to generalize across clients \cite{karimireddy2019scaffold, wang2020tackling}.

Traditional FL methods, such as FedAvg \cite{MMRH2017}, rely on a central server to aggregate local model updates into a global model through parameter averaging. While effective when data is similarly distributed, FedAvg struggles with non-IID data, as local optima diverge from the global target, leading to poor performance on individual clients \cite{zhao2018federated}. Conversely, relying solely on individual learning without FL collaboration leads to poor generalization as well, as clients lack sufficient data to train robust models independently \cite{chen2021bridging}.
To overcome this, personalized FL has gained attention, aiming to tailor models to each client’s unique data while still benefiting from collaborative learning \cite{li2021ditto}. Personalized FL seeks to strike a balance between global knowledge sharing and local adaptation, a task complicated by the need to maintain privacy and efficiency in heterogeneous settings \cite{tan2022towards}.

Numerous strategies have been developed to enable personalization in FL. For example, local customization techniques, such as FedProx \cite{li2018fedprox}, modify FedAvg by introducing regularization to align local updates with the global model. While FedPer \cite{arivazhagan2019fedper} splits neural networks into shared base layers trained server-side and personalized layers fine-tuned by clients. Bi-level optimization offers another avenue, exemplified by pFedMe \cite{dinh2020pfedme}, which separates global model aggregation from local personalization, using the global model as a reference point to guide client-specific solutions—an approach that proves effective but less so with limited data. Meta-learning strategies, like Per-FedAvg \cite{fallah2020personalized}, draw from Model-Agnostic Meta-Learning to craft an initial shared model that clients can quickly adapt, though computational challenges arise from Hessian approximations \cite{jiang2019improving}. Multi-task learning contributes frameworks like MOCHA \cite{smith2017multitask}, which tackles both statistical and system-level diversity by treating clients as distinct tasks. Meanwhile, methods like APFL \cite{deng2020adaptive} blend local and global models adaptively, balancing personalization with collaboration. Clustering-based approaches, such as those proposed by \cite{ghosh2020efficient}, group clients with similar data distributions to improve personalization efficiency. Despite these advances, many PFL approaches grapple with overfitting when client datasets are small, underscoring the need for robust solutions tailored to data scarcity \cite{zhang2021survey}.

Knowledge Distillation (KD), introduced by Hinton et al. \cite{hinton2015distilling}, provides another powerful tool to enhance FL under heterogeneity. KD transfers knowledge from a complex teacher model to a simpler student model using a weighted combination of loss terms, enabling the student to learn both from ground-truth labels and the teacher’s broader predictive patterns. In FL, KD has been adapted to distill global knowledge into local models, improving convergence and robustness to non-IID data \cite{lin2020ensemble}. Recent works have built on this idea in various ways. For instance, Jeong and Kountouris \cite{jeong2023kdpfl} propose KD-PDFL, a decentralized personalized FL method that uses KD to align local models based on estimated similarities, focusing on personalization over global coherence. Seo et al. \cite{seo2022federated} introduce Federated Distillation (FD), which reduces communication costs by sharing model outputs instead of parameters, though it sacrifices accuracy in highly non-IID scenarios \cite{zhu2021data}. Yao et al. \cite{yao2023fedgkd} present FedGKD, which regularizes local training with an ensemble of historical global models via KD, achieving strong performance but relying on a centralized server and increased communication.

We propose a novel FL algorithm that integrates KD with a weighted combination loss to tackle non-IID challenges in a centralized setting. Our approach draws from the insight of bi-level optimization separating global and local optimization, using the global model as a teacher to guide local models via KD, while optimizing both levels concurrently. Unlike FD, which prioritizes communication efficiency, or KD-PDFL, which emphasizes personalization, our method balances global convergence and local regularization. Compared to FedGKD, we streamline the process by focusing on the current global model rather than historical ensembles, enhancing efficiency and robustness. We evaluate our algorithm on the MNIST dataset and a synthetic dataset under non-IID conditions, demonstrating improved accuracy and convergence speed over other state-of-the-art (SOTA) algorithms.

The remainder of this manuscript is organized as follows: Section 2 details the problem formulation and our proposed algorithm. Section 3 presents experimental results, and Section 4 concludes with future research directions.
\section{Materials and Methods} 
In this section, we outline the problem formulation and introduce our proposed FL algorithm that integrates KD with a weighted combination loss to address data heterogeneity. We begin by describing the conventional FL model, followed by an overview of KD as a foundational technique, and conclude with the details of our proposed method.

\subsection{Conventional FL Model}
\label{subsec:conventional_fl}

In conventional FL, a system comprises a central server and \( N \) clients, each holding a local dataset \( \mathcal{D}_i \) drawn from a potentially distinct distribution \( P_i \), reflecting the non-IID nature of real-world scenarios. The objective is to collaboratively train a global model parameterized by \( w \in \mathbb{R}^d \) without sharing raw data. This is typically formulated as an optimization problem:

\begin{equation}
\min_{w \in \mathbb{R}^d} \left\{ f(w) := \frac{1}{N} \sum_{i=1}^N f_i(w) \right\},
\label{eq:conventional_fl}
\end{equation}
where \( f_i(w) = \mathbb{E}_{\mathcal{D}_i \sim P_i} [\tilde{f}_i(w; \mathcal{D}_i)] \) represents the expected loss over client \( i \)’s data distribution, and \( \tilde{f}_i(w; \mathcal{D}_i) \) is the loss for a specific data sample \( \mathcal{D}_i \). A widely adopted approach to solve this is FedAvg \cite{MMRH2017}, where at each round \( t \):
1. The server broadcasts the global model \( w_t \) to a subset of clients.
2. Each selected client \( i \) performs local updates (e.g., via stochastic gradient descent) on \( w_t \) using \( \mathcal{D}_i \) to obtain a local model \( w_{i,t+1} \).
3. The server aggregates these updates as \( w_{t+1} = \sum_{i \in \mathcal{S}_t} \frac{|\mathcal{D}_i|}{\sum_{j \in \mathcal{S}_t} |\mathcal{D}_j|} w_{i,t+1} \), where \( \mathcal{S}_t \) is the sampled client subset. While effective under IID conditions, FedAvg struggles with non-IID data, as local objectives diverge, leading to client drift and poor generalization \cite{zhao2018federated}.

\subsection{Knowledge Distillation}
\label{subsec:kd}

Knowledge Distillation (KD), introduced by \cite{hinton2015distilling}, is a technique to transfer knowledge from a pre-trained teacher model to a student model. In its original form, the student is trained to minimize a weighted combination of two loss terms: the cross-entropy (CE) loss with hard labels and the Kullback-Leibler (KL) divergence between the teacher’s and student’s softened predictions (logits). For a student model with parameters \( \theta \) and a teacher model with parameters \( w \), the KD loss is:

\begin{equation}
\mathcal{L}_{\text{KD}}(\theta) = (1 - \alpha) \cdot \mathcal{L}_{\text{CE}}(\theta, y) + \alpha \cdot T^2 \cdot \text{KL} \left( \frac{p_w}{T}, \frac{p_\theta}{T} \right),
\label{eq:kd_loss}
\end{equation}
where \( \mathcal{L}_{\text{CE}} \) is the CE loss with true labels \( y \), \( p_w \) and \( p_\theta \) are the teacher’s and student’s output probabilities, \( T \) is a temperature parameter softening the logits, and \( \alpha \in [0, 1] \) balances the two terms. The factor \( T^2 \) adjusts the KL divergence magnitude due to temperature scaling. KD enables the student to inherit the teacher’s generalization capabilities, making it a promising tool for regularizing local models in FL under heterogeneity.

\subsection{Proposed Method}
\label{subsec:proposed_method}

We propose a novel FL algorithm, termed pFedKD-WCL (Personalized Federated Knowledge Distillation with Weighted Combination Loss), to balance global convergence and local personalization in non-IID settings. Our method decouples global and local optimization, using the global model as a teacher to guide local updates via KD, while refining the global model with insights from local data.

\subsubsection{Problem Formulation}

We formulate pFedKD-WCL as a bi-level optimization problem. At the client level, each client \( i \) optimizes a personalized model \( \theta_i \in \mathbb{R}^d \) using a KD-based loss with the global model \( w \) as the teacher:

\begin{equation}
\min_{\theta_i \in \mathbb{R}^d} \left\{ F_i(\theta_i, w) := (1 - \alpha) \cdot f_i(\theta_i) + \alpha \cdot \text{KL}(p_w, p_{\theta_i}) \right\},
\label{eq:local_objective}
\end{equation}

where \( f_i(\theta_i) = \mathbb{E}_{\mathcal{D}_i \sim P_i} [\tilde{f}_i(\theta_i; \mathcal{D}_i)] \) is the local CE loss, and \( \text{KL}(p_w, p_{\theta_i}) \) measures the divergence between the global model’s predictions \( p_w \) and the local model’s predictions \( p_{\theta_i} \) on \( \mathcal{D}_i \), with logits softened by temperature \( T \) (set to 1 for simplicity). At the server level, the global model \( w \) is optimized to minimize the average divergence from local models:

\begin{equation}
\min_{w \in \mathbb{R}^d} \left\{ F(w) := \frac{1}{N} \sum_{i=1}^N \text{KL}(p_{\hat{\theta}_i(w)}, p_w) \right\},
\label{eq:global_objective}
\end{equation}

where \( \hat{\theta}_i(w) = \arg\min_{\theta_i} F_i(\theta_i, w) \) is the optimal local model for client \( i \) given \( w \). This formulation encourages the global model to align with local predictions, enhancing robustness to heterogeneity.

\subsubsection{Algorithm Description}
Our algorithm operates in rounds, as outlined in Algorithm~\ref{alg:fedkd_wcl}:
\begin{algorithm}
\caption{pFedKD-WCL Algorithm}
\label{alg:fedkd_wcl}
\begin{algorithmic}[1]
\STATE \textbf{Input:} \( T \) (rounds), \( R \) (local steps), \( S \) (clients per round), \( \eta \) (learning rate), \( \alpha \) (KD weight), \( w_0 \) (initial global model)
\FOR{\( t = 0 \) to \( T-1 \)}
    \STATE Server samples subset \( \mathcal{S}_t \) of \( S \) clients
    \STATE Server broadcasts \( w_t \) to all clients in \( \mathcal{S}_t \)
    \FOR{each client in \( \mathcal{S}_t \) in parallel}
        \FOR{\( r = 0 \) to \( R-1 \)}
            \STATE Sample mini-batch \( \mathcal{D}_{i,r}^t \) from \( \mathcal{D}_i \)
            \STATE Compute loss \( F_i(\theta_{i,r}^t, w_t) \) using Eq.~\eqref{eq:local_objective}
            \STATE Update \( \theta_{i,r+1}^t = \theta_{i,r}^t - \eta \nabla_{\theta_i} F_i(\theta_{i,r}^t, w_t) \)
        \ENDFOR
        \STATE Set \( \hat{\theta}_i^t = \theta_{i,R}^t \)
    \ENDFOR
    \STATE Clients in \( \mathcal{S}_t \) send \( \hat{\theta}_i^t \) to server
    \STATE Server updates \( w_{t+1} = w_t - \eta \frac{1}{S} \sum_{i \in \mathcal{S}_t} \nabla_w \text{KL}(p_{\hat{\theta}_i^t}, p_{w_t}) \) with Eq.~\eqref{eq:global_objective}
\ENDFOR
\STATE \textbf{Output:} Global model \( w_T \), personalized models \( \{ \hat{\theta}_i^T \} \)
\end{algorithmic}
\end{algorithm}
Unlike FedAvg, which aggregates parameters, pFedKD-WCL leverages KD to transfer knowledge bidirectionally: the global model regularizes local training, and local predictions refine the global model. The parameter \( \alpha \) controls the trade-off between local fit and global alignment, with \( \alpha = 0 \) reducing to standalone local training and \( \alpha = 1 \) prioritizing global consistency. Compared to pFedMe, which uses an \( l_2 \)-norm penalty, our KD-based approach captures richer predictive relationships, potentially improving generalization on non-IID data.

\section{Experiments and Results}

To evaluate the performance of our proposed method, we conduct a series of experiments comparing it against four established federated learning algorithms: FedAvg, FedProx, PerFedAvg and pFedMe. These experiments are performed on two datasets---MNIST and a synthetic dataset---both configured with non-IID data distributions to simulate realistic federated learning scenarios with heterogeneous client data. In the following subsections, we describe the experimental settings, including the datasets, their non-IID partitioning, and the model architectures used, followed by a detailed discussion of the hyperparameter configurations for all methods.

\subsection{Experimental Setting}

The experiments utilize two datasets: MNIST and a synthetic dataset. The MNIST dataset, introduced by LeCun et al.\cite{lecun1998gradient}, comprises 70,000 grayscale images of handwritten digits (0--9), each $28\times28$ pixels, with 60,000 training and 10,000 testing samples. To create a non-IID setting, we partition the training data across 20 clients by clustering samples based on digit labels, ensuring each client predominantly holds data from a subset of digits (e.g., 2 digits per client) with varying sample sizes to reflect imbalance. This partitioning follows a Dirichlet distribution with a concentration parameter $\alpha = 0.5$, as suggested by Hsu et al.\cite{hsu2019}, resulting in a challenging heterogeneous distribution. The synthetic dataset is generated to test generalization under controlled heterogeneity, following the methodology of Li et al.\cite{li2018fedprox}. It consists of 100 clients, each with varied samples of 60-dimensional feature vectors and 10 classes offering a simplified yet diverse counterpart to the MNIST dataset for evaluating model robustness. Features are drawn from a Gaussian distribution $\mathcal{N}(\mu_c, \sigma_c^2)$, where $\mu_c$ and $\sigma_c$ vary per client and class, and labels are assigned using a logistic classifier with client-specific weights, creating diverse decision boundaries across clients. The dataset of each client is partitioned into training and test sets, with 75\% of the samples allocated to training and 25\% to testing, using random permutation to ensure an unbiased split, and the data are converted to PyTorch tensors for compatibility with deep learning frameworks. 

Two models are employed to assess performance across varying complexities. The multinomial logistic regression (MLR) model is a simple linear classifier with an input layer of 784 units (for MNIST or 60 units for synthetic input) and an output layer of 10 classes, followed by a log-softmax activation to produce log-probabilities. The multilayer perceptron (MLP) is a neural network with an input layer of 784 units (for MNIST or 60 units for synthetic input), a hidden layer of 128 units with ReLU activation, and an output layer of 10 units with log-softmax activation. The MLP's additional capacity allows it to capture more complex patterns, while the MLR provides a lightweight baseline. Both models are trained to minimize the negative log-likelihood (NLL) loss, consistent with their log-softmax outputs and standard classification objectives in federated learning.

We perform all experiments in this paper using a server node with 4 NVIDIA V100 GPUs with 16GB memory, 4 Intel(R) Xeon(R) CPUs with 12 cores @ 2.40GHz) and 360 GB memory.

\subsection{Experimental Hyperparameter Settings}
To ensure a fair and comprehensive comparison, we carefully configure hyperparameters for each method while maintaining a consistent experimental framework. All methods share a baseline setting: 20 clients participate in the federation (100 in synthetic data), 5 of which are selected per round (10 in synthetic data), balancing communication efficiency and model updates. Local training uses a batch size of 20 and runs for 20 local epochs per round, with a learning rate of 0.01 applied via stochastic gradient descent (SGD) without momentum.

For FedAvg, these common hyperparameters suffice, as it relies solely on local SGD and global averaging without additional tuning parameters. FedProx introduces a proximal term weight ($\mu$) set to 0.01, which controls the regularization strength towards the global model. PerFedAvg incorporates a beta parameter ($\beta$) set to 0.002. $\lambda$ in pFedMe varies in different settings, and was shown in Table~\ref{tab1}. Our pFedKD method uses a knowledge distillation weight ($\alpha$) of 0.1, weighting the KL-divergence loss between local and global model predictions.

All models are initialized randomly, and client data is pre-partitioned to ensure identical non-IID distributions across methods. Evaluation metrics differ by algorithm design: FedAvg and FedProx report global model accuracy and loss, while PerFedAvg, pFedMe and pFedKD-WCL provide personalized metrics after local adaptation, reflecting their focus on client-specific performance. 

\subsection{Effect of the hyperparameter}
\begin{figure}
\centering
\includegraphics[width=14cm]{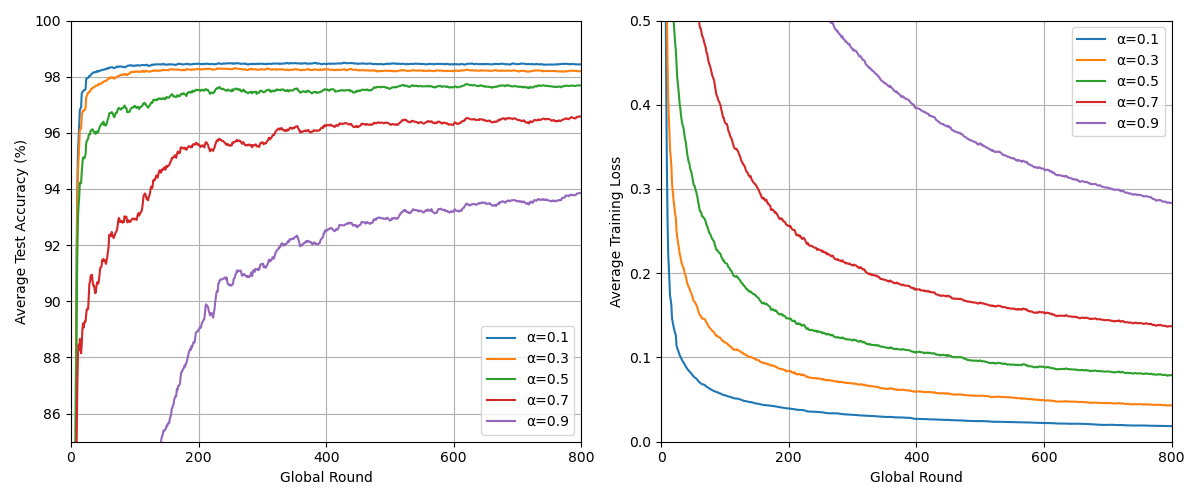}
\caption{Effect of KD weight \( \alpha \) on the MLR model using the MNIST dataset. Left: Average test accuracy (\%) over 800 global rounds, Right: Average training loss.\label{fig:alpha_mlr}}
\end{figure}

\begin{figure}
\centering
\includegraphics[width=14cm]{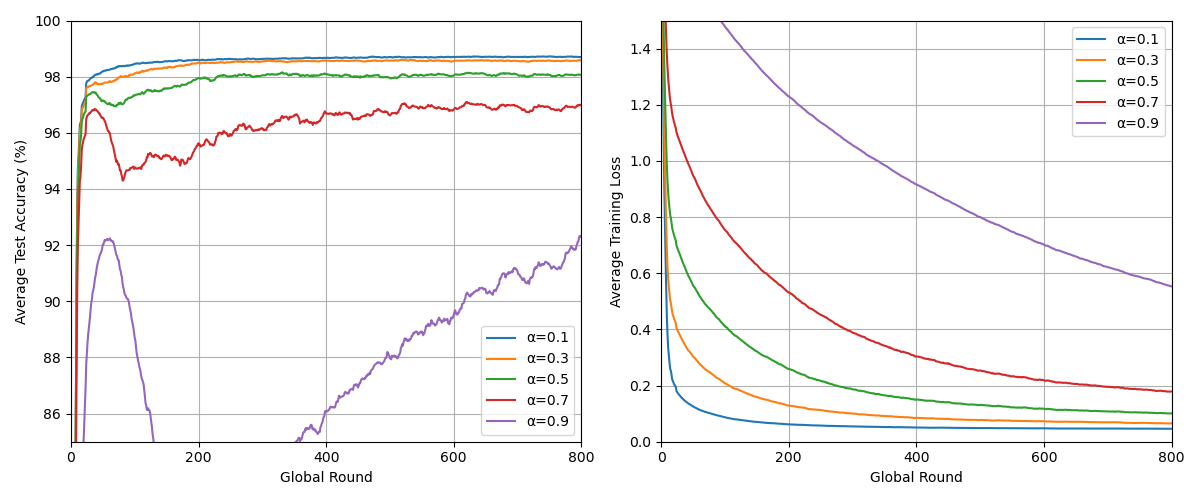}
\caption{Effect of KD weight \( \alpha \) on the MLP model using the MNIST dataset. Left: Average test accuracy (\%) over 800 global rounds. Right: Average training loss.\label{fig:alpha_mlp}}
\end{figure}

To evaluate the effect of the KD weight parameter \( \alpha \) in our algorithm, we conducted experiments on the MNIST dataset with two models: a MLR model (Figure~\ref{fig:alpha_mlr}) and a MLP model (Figure~\ref{fig:alpha_mlp}) with $ \alpha = 0.1, 0.3, 0.5, 0.7, 0.9$. For the MLR model, the left plot shows that average test accuracy stabilizes above 98\% within 200 rounds, with \( \alpha = 0.1 \) and \( \alpha = 0.3 \), while \( \alpha = 0.9 \) fluctuates and plateaus around 94\%. The corresponding training loss plot indicates that \( \alpha = 0.1 \) and \( \alpha = 0.3 \) converge to below 0.1 after 200 rounds, whereas \( \alpha = 0.9 \) retains a higher loss of approximately 0.28. In contrast, the MLP model exhibits more pronounced variability: accuracy initially surges but drops sharply for \( \alpha = 0.9 \) (below 90\% after 100 rounds) before gradually recovering to around 92\% by 800 rounds, while \( \alpha = 0.1 \) and \( \alpha = 0.3 \) maintain around 98\% with slight fluctuations. The MLP’s training loss decreases steadily for all \( \alpha \) values, but \( \alpha = 0.9 \) remains elevated (above 0.5), compared to lower losses (below 0.2) for \( \alpha = 0.1 \) and \( \alpha = 0.3 \). This behavior indicates that with high \( \alpha \), the MLP may over-adjust its weights to mimic the global model, ignoring local data patterns, which exacerbates the accuracy drop compared to the simpler MLR model. The MLP’s deeper architecture makes it more sensitive to the choice of \( \alpha \), highlighting the need for careful tuning in non-IID settings. The results suggest that for deeper models like MLPs in non-IID settings, a lower \( \alpha \) (e.g., 0.1 or 0.3) strikes a better balance between local adaptation and global regularization, avoiding the instability seen with high \( \alpha \). These findings underscore the importance of model-specific tuning to mitigate instability in complex architectures under heterogeneous data distributions.

\subsection{Performance Comparison Results}
\begin{figure}
\centering
\includegraphics[width=14cm]{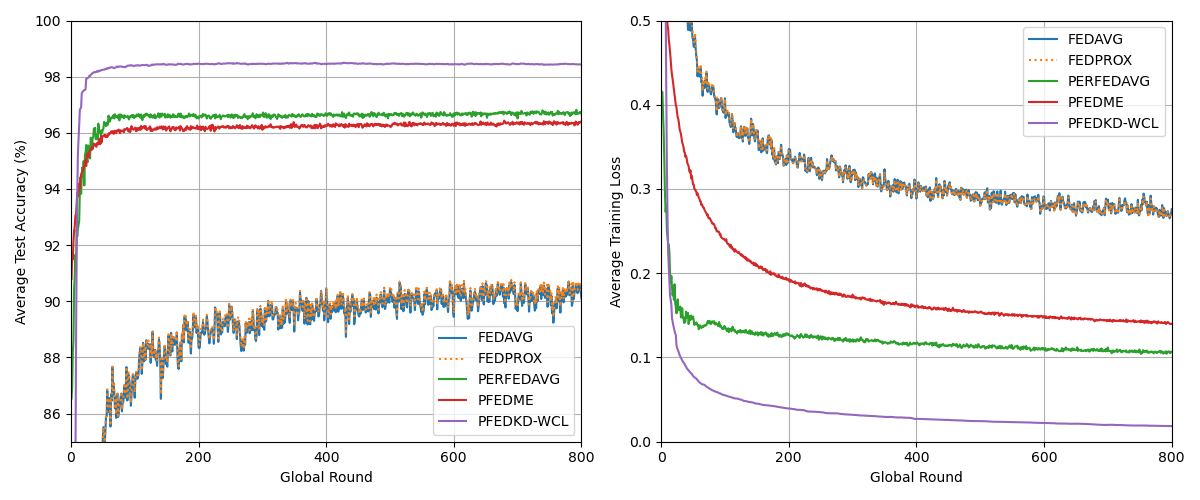}
\caption{Performance comparisons of the average test accuracy and training loss of different algorithms on the MNIST dataset with multinomial logistic regression (MLR).\label{fig3}}
\end{figure}

\begin{figure}
\centering
\includegraphics[width=14 cm]{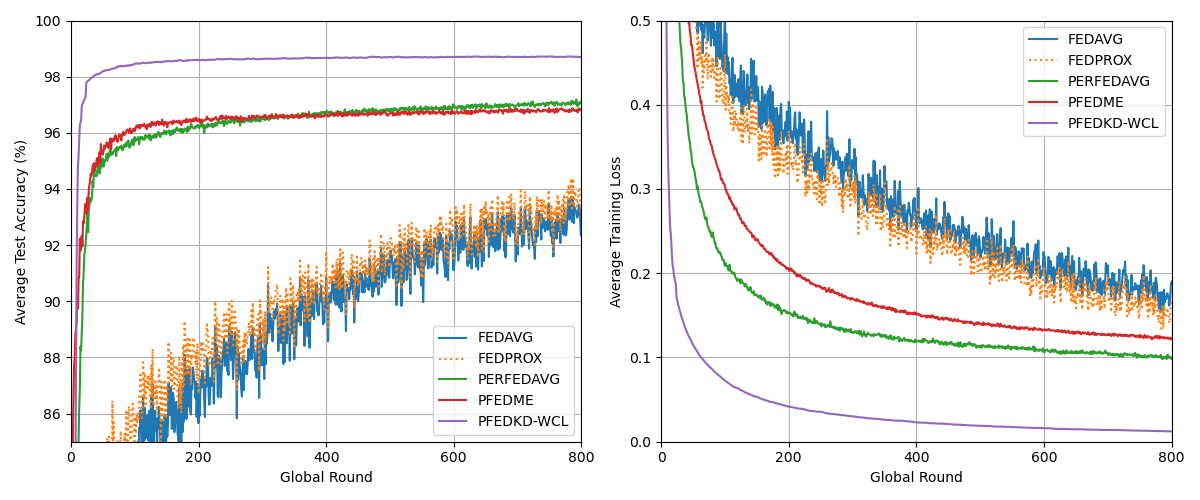}
\caption{Performance comparisons of the average test accuracy and training loss of different algorithms on the MNIST dataset with two-layer multilayer perceptron (MLP).\label{fig4}}
\end{figure}

\begin{figure}
\centering
\includegraphics[width=14cm]{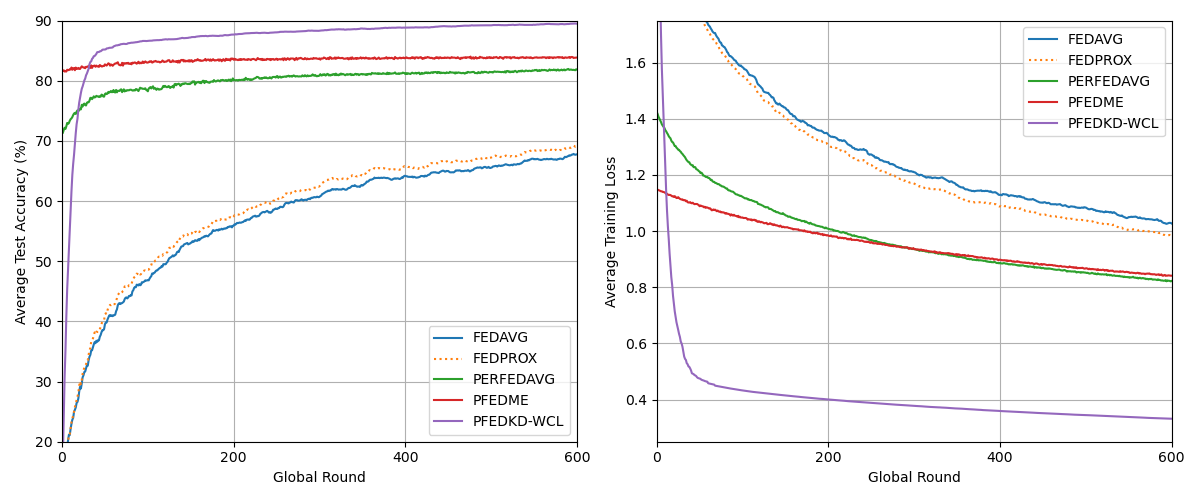}
\caption{Performance comparisons of the average test accuracy and training loss of different algorithms on the synthetic dataset with multinomial logistic regression (MLR).\label{fig5}}
\end{figure}

\begin{figure}
\centering
\includegraphics[width=14 cm]{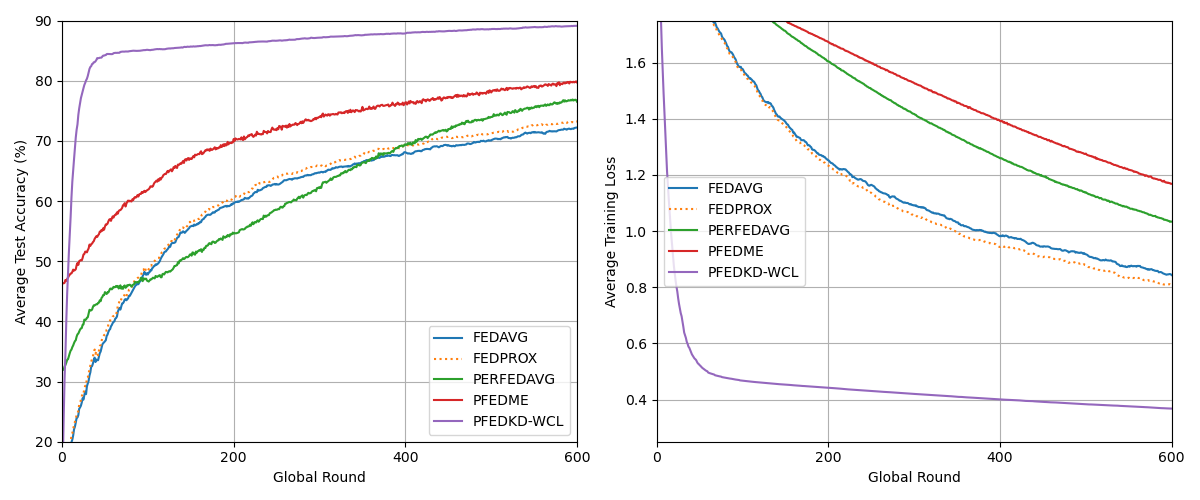}
\caption{Performance comparisons of the average test accuracy and training loss of different algorithms on the synthetic dataset with two-layer multilayer perceptron (MLP).\label{fig6}}
\end{figure}

Figures~\ref{fig3} and~\ref{fig4} illustrate the performance comparison of SOTA FL algorithms on the MNIST dataset using MLR and MLP models, respectively. Table~\ref{tab1} presents the average test accuracy after 800 training rounds for MNIST and 600 rounds for synthetic, with fine-tuned hyperparameters. In terms of prediction accuracy in MNIST dataset, pFedKD-WCL demonstrates improvements of 9.2\%, 9.1\%, 1.8\%, and 2.1\% over FedAvg, FedProx, Per-FedAvg, and pFedMe respectively with the MLR model, and 6.8\%, 6.6\%, 1.6\%, and 1.9\% with the MLP model. Figures~\ref{fig5} and~\ref{fig6} depict the performance of these algorithms on a synthetic dataset using MLR and MLP models, respectively. With the MLR model, pFedKD-WCL outperforms FedAvg, FedProx, Per-FedAvg, and pFedMe by 32.0\%, 30.0\%, 7.6\%, and 6.7\%, respectively. Similarly, with the MLP model, the improvements are 23.4\%, 21.7\%, 12.6\%, and 11.6\% compared to the same baseline methods.

\begin{table}
\caption{Performance comparisons of the average test accuracy. \label{tab1}}
\begin{tabular}{lccccc}
        \toprule
        \multirow{2}{*}{Algorithm} & \multirow{2}{*}{Model} & \multicolumn{2}{c}{MNIST} & \multicolumn{2}{c}{Synthetic} \\
        \cmidrule(lr){3-4} \cmidrule(lr){5-6}
        & & Hyperparameters  & Accuracy (\%) & Hyperparameters  & Accuracy (\%) \\
        \midrule
        FedAvg & MLR & -  & $90.10 \pm 0.10$ & - &  $67.76 \pm 0.09$  \\
        FedProx & MLR & $\mu=0.01$  & $90.23 \pm 0.09$ & $\mu=0.01$ & $69.11 \pm 0.02$  \\
        Per-FedAvg & MLR & -  & $96.68 \pm 0.02$ & - & $81.86 \pm 0.03$  \\
        pFedMe & MLR & $\lambda=15$  & $96.41 \pm 0.01$ & $\lambda=20$   &  $83.84 \pm 0.02$ \\
        pFedKD-WDL & MLR & $\alpha=0.1$  & $98.44 \pm 0.01$ & $\alpha=0.1$   & $89.48 \pm 0.01$ \\
        \midrule
        FedAvg & MLP & -  & $92.41 \pm 0.10$ & - & $72.22 \pm 0.09$  \\
        FedProx & MLP & $\mu=0.01$  & $92.64 \pm 0.10$ & $\mu=0.01$ & $73.22 \pm 0.09$  \\
        Per-FedAvg & MLP & -  & $97.13 \pm 0.01$ & - & $76.51 \pm 0.10$  \\
        pFedMe & MLP & $\lambda=15$  & $96.84 \pm 0.01$ & $\lambda=30$   & $79.71 \pm 0.10$  \\
        pFedKD-WDL & MLP & $\alpha=0.1$  & $98.71 \pm 0.01$ & $\alpha=0.1$   & $89.12 \pm 0.02$  \\
        \bottomrule
\end{tabular}
\end{table}


\section{Conclusions}
This study has demonstrated the effectiveness of our proposed FedKD-WCL algorithm in addressing the challenges posed by statistical heterogeneity in FL, particularly in non-IID data settings. By integrating knowledge distillation with a bi-level optimization framework, FedKD-WCL successfully balances global model convergence and local personalization, outperforming established baseline methods across diverse datasets. The experimental results underscore the advantage of using the global model as a teacher to guide local training, with deeper models like multilayer perceptrons showing sensitivity to the KD weight parameter, highlighting the importance of model-specific tuning in non-IID environments. While these findings affirm the robustness of FedKD-WCL, its reliance on centralized coordination and potential computational demands present areas for improvement, especially in large-scale deployments. Future research could explore adaptive tuning strategies for the KD weight, thereby enhancing the algorithm’s applicability to real-world FL scenarios.

\bibliographystyle{unsrt}  
 \bibliography{ref}

\begin{thebibliography}{10}

\bibitem{MMRH2017}
Brendan McMahan, Eider Moore, Daniel Ramage, Seth Hampson, and Blaise~Aguera y~Arcas.
\newblock Communication-efficient learning of deep networks from decentralized data.
\newblock In {\em Artificial intelligence and statistics}, pages 1273--1282. PMLR, 2017.

\bibitem{kairouz2019advances}
Peter Kairouz, H.~Brendan McMahan, Brendan Avent, Aurélien Bellet, Mehdi Bennis, Arjun Nitin~Bhagoji, Kallista Bonawitz, Zachary Charles, Graham Cormode, Rachel Cummings, Rafael G.~L. D’Oliveira, Hubert Eichner, Salim El~Rouayheb, David Evans, Josh Gardner, Zachary Garrett, Adrià Gascón, Badih Ghazi, Phillip~B. Gibbons, Marco Gruteser, Zaid Harchaoui, Chaoyang He, Lie He, Zhouyuan Huo, Ben Hutchinson, Justin Hsu, Martin Jaggi, Tara Javidi, Gauri Joshi, Mikhail Khodak, Jakub Konecný, Aleksandra Korolova, Farinaz Koushanfar, Sanmi Koyejo, Tancrède Lepoint, Yang Liu, Prateek Mittal, Mehryar Mohri, Richard Nock, Ayfer Özgür, Rasmus Pagh, Hang Qi, Daniel Ramage, Ramesh Raskar, Mariana Raykova, Dawn Song, Weikang Song, Sebastian~U. Stich, Ziteng Sun, Ananda~Theertha Suresh, Florian Tramèr, Praneeth Vepakomma, Jianyu Wang, Li~Xiong, Zheng Xu, Qiang Yang, Felix~X. Yu, Han Yu, and Sen Zhao.
\newblock Advances and open problems in federated learning.
\newblock {\em Foundations and Trends® in Machine Learning}, 14(1–2):1–210, June 2021.

\bibitem{yang2019federated}
Qiang Yang, Yang Liu, Tianjian Chen, and Yongxin Tong.
\newblock Federated machine learning: Concept and applications.
\newblock {\em ACM Transactions on Intelligent Systems and Technology (TIST)}, 10(2):1--19, 2019.

\bibitem{li2019convergence}
Tian Li, Anit~Kumar Sahu, Manzil Zaheer, Maziar Sanjabi, Ameet Talwalkar, and Virginia Smith.
\newblock Federated optimization in heterogeneous networks.
\newblock {\em Proceedings of Machine learning and systems}, 2:429--450, 2020.

\bibitem{sattler2019robust}
Felix Sattler, Simon Wiedemann, Klaus-Robert M{\"u}ller, and Wojciech Samek.
\newblock Robust and communication-efficient federated learning from non-iid data.
\newblock {\em IEEE transactions on neural networks and learning systems}, 31(9):3400--3413, 2019.

\bibitem{karimireddy2019scaffold}
Sai~Praneeth Karimireddy, Satyen Kale, Mehryar Mohri, Sashank Reddi, Sebastian Stich, and Ananda~Theertha Suresh.
\newblock Scaffold: Stochastic controlled averaging for federated learning.
\newblock In {\em International conference on machine learning}, pages 5132--5143. PMLR, 2020.

\bibitem{wang2020tackling}
Jianyu Wang, Qinghua Liu, Hao Liang, Gauri Joshi, and H~Vincent Poor.
\newblock Tackling the objective inconsistency problem in heterogeneous federated optimization.
\newblock {\em Advances in neural information processing systems}, 33:7611--7623, 2020.

\bibitem{zhao2018federated}
Yue Zhao, Meng Li, Liangzhen Lai, Naveen Suda, Damon Civin, and Vikas Chandra.
\newblock Federated learning with non-iid data.
\newblock {\em arXiv preprint arXiv:1806.00582}, 2018.

\bibitem{chen2021bridging}
Hong-You Chen and Wei-Lun Chao.
\newblock On bridging generic and personalized federated learning for image classification.
\newblock {\em arXiv preprint arXiv:2107.00778}, 2021.

\bibitem{li2021ditto}
Tian Li, Shengyuan Hu, Ahmad Beirami, and Virginia Smith.
\newblock Ditto: Fair and robust federated learning through personalization.
\newblock In {\em International conference on machine learning}, pages 6357--6368. PMLR, 2021.

\bibitem{tan2022towards}
Alysa~Ziying Tan, Han Yu, Lizhen Cui, and Qiang Yang.
\newblock Towards personalized federated learning.
\newblock {\em IEEE transactions on neural networks and learning systems}, 34(12):9587--9603, 2022.

\bibitem{li2018fedprox}
Tian Li, Anit~Kumar Sahu, Manzil Zaheer, Maziar Sanjabi, Ameet Talwalkar, and Virginia Smith.
\newblock Federated optimization in heterogeneous networks.
\newblock {\em Proceedings of Machine learning and systems}, 2:429--450, 2020.

\bibitem{arivazhagan2019fedper}
Manoj~Ghuhan Arivazhagan, Vinay Aggarwal, Aaditya~Kumar Singh, and Sunav Choudhary.
\newblock Federated learning with personalization layers.
\newblock {\em arXiv preprint arXiv:1912.00818}, 2019.

\bibitem{dinh2020pfedme}
Canh T~Dinh, Nguyen Tran, and Josh Nguyen.
\newblock Personalized federated learning with moreau envelopes.
\newblock {\em Advances in neural information processing systems}, 33:21394--21405, 2020.

\bibitem{fallah2020personalized}
Alireza Fallah, Aryan Mokhtari, and Asuman Ozdaglar.
\newblock Personalized federated learning: A meta-learning approach.
\newblock {\em arXiv preprint arXiv:2002.07948}, 2020.

\bibitem{jiang2019improving}
Yihan Jiang, Jakub Kone{\v{c}}n{\`y}, Keith Rush, and Sreeram Kannan.
\newblock Improving federated learning personalization via model agnostic meta learning.
\newblock {\em arXiv preprint arXiv:1909.12488}, 2019.

\bibitem{smith2017multitask}
Virginia Smith, Chao-Kai Chiang, Maziar Sanjabi, and Ameet~S Talwalkar.
\newblock Federated multi-task learning.
\newblock {\em Advances in neural information processing systems}, 30, 2017.

\bibitem{deng2020adaptive}
Yuyang Deng, Mohammad~Mahdi Kamani, and Mehrdad Mahdavi.
\newblock Adaptive personalized federated learning.
\newblock {\em arXiv preprint arXiv:2003.13461}, 2020.

\bibitem{ghosh2020efficient}
Avishek Ghosh, Jichan Chung, Dong Yin, and Kannan Ramchandran.
\newblock An efficient framework for clustered federated learning.
\newblock {\em Advances in neural information processing systems}, 33:19586--19597, 2020.

\bibitem{zhang2021survey}
Chen Zhang, Yu~Xie, Hang Bai, Bin Yu, Weihong Li, and Yuan Gao.
\newblock A survey on federated learning.
\newblock {\em Knowledge-Based Systems}, 216:106775, 2021.

\bibitem{hinton2015distilling}
Geoffrey Hinton, Oriol Vinyals, and Jeff Dean.
\newblock Distilling the knowledge in a neural network.
\newblock {\em arXiv preprint arXiv:1503.02531}, 2015.

\bibitem{lin2020ensemble}
Tao Lin, Lingjing Kong, Sebastian~U Stich, and Martin Jaggi.
\newblock Ensemble distillation for robust model fusion in federated learning.
\newblock {\em Advances in neural information processing systems}, 33:2351--2363, 2020.

\bibitem{jeong2023kdpfl}
Eunjeong Jeong and Marios Kountouris.
\newblock Personalized decentralized federated learning with knowledge distillation.
\newblock In {\em ICC 2023-IEEE International Conference on Communications}, pages 1982--1987. IEEE, 2023.

\bibitem{seo2022federated}
Hyowoon Seo, Jihong Park, Seungeun Oh, Mehdi Bennis, and Seong-Lyun Kim.
\newblock 16 federated knowledge distillation.
\newblock {\em Machine Learning and Wireless Communications}, 457, 2022.

\bibitem{zhu2021data}
Zhuangdi Zhu, Junyuan Hong, and Jiayu Zhou.
\newblock Data-free knowledge distillation for heterogeneous federated learning.
\newblock In {\em International conference on machine learning}, pages 12878--12889. PMLR, 2021.

\bibitem{yao2023fedgkd}
Dezhong Yao, Wanning Pan, Yutong Dai, Yao Wan, Xiaofeng Ding, Chen Yu, Hai Jin, Zheng Xu, and Lichao Sun.
\newblock Fedgkd: Toward heterogeneous federated learning via global knowledge distillation.
\newblock {\em IEEE Transactions on Computers}, 73(1):3--17, 2023.

\bibitem{lecun1998gradient}
Yann LeCun, L{\'e}on Bottou, Yoshua Bengio, and Patrick Haffner.
\newblock Gradient-based learning applied to document recognition.
\newblock {\em Proceedings of the IEEE}, 86(11):2278--2324, 1998.

\bibitem{hsu2019}
Tzu-Ming~Harry Hsu, Hang Qi, and Matthew Brown.
\newblock Measuring the effects of non-identical data distribution for federated visual classification, 2019.
\newblock arXiv:1909.06335.

\end{thebibliography}


\end{document}